\documentclass[lettersize,journal]{IEEEtran}
\usepackage{amsmath,amsfonts}
\usepackage{algorithmic}
\usepackage{array}
\usepackage[caption=false,font=normalsize,labelfont=sf,textfont=sf]{subfig}
\usepackage{textcomp}
\usepackage{stfloats}
\usepackage{url}
\usepackage{verbatim}
\usepackage{graphicx}
\usepackage{bm} % Added for bold math symbols like \boldsymbol{\epsilon}
\usepackage{multirow}
\usepackage{tabularx}
\usepackage{xcolor}
\def\BibTeX{{\rm B\kern-.05em{\sc i\kern-.025em b}\kern-.08em
    T\kern-.1667em\lower.7ex\hbox{E}\kern-.125emX}}
\usepackage{balance}

\begin{document}

\title{MISTY: High-Throughput Motion Planning via Mixer-based Single-step Drifting}

\author{Yining Xing$^{1}$, Zehong Ke$^{1}$, Yiqian Tu$^{1}$, Zhiyuan Liu$^{1}$, Wenhao Yu$^{2,*}$, and Jianqiang Wang$^{1}$
\thanks{$^{1}$Yining Xing, Zehong Ke, Yiqian Tu, Zhiyuan Liu and Jianqiang Wang are with the School of Vehicle and Mobility, Tsinghua University, Beijing, China.}
\thanks{$^{2}$Wenhao Yu is with the State Key Laboratory of Intelligent Green Vehicle and Mobility, Tsinghua University, Beijing, China (Corresponding author, e-mail: eyre530056@gmail.com).}
\thanks{This work was supported by National Key R\&D Program of China: 2023YFB2504400, National Natural Science Foundation of China grant 52572477, and Independent Research Project of the State Key Laboratory of Intelligent Green Vehicle and Mobility, Tsinghua University (No. ZZ-PY-20250409).}}

% \author{Anonymous Author(s)
% \thanks{This paper is under double-anonymous review. Author identities, affiliations, and acknowledgments are hidden for the review process.}}

\markboth{IEEE Robotics and Automation Letters,~Vol.~X, No.~X, Month~2026}%
{Author \MakeLowercase{\textit{et al.}}: MISTY: High-Throughput Motion Planning}

\maketitle

\begin{abstract}
Multi-modal trajectory generation is essential for safe autonomous driving, yet existing diffusion-based planners suffer from high inference latency due to iterative neural function evaluations. This paper presents MISTY (Mixer-based Inference for Single-step Trajectory-drifting Yield), a high-throughput generative motion planner that achieves state-of-the-art closed-loop performance with pure single-step inference. MISTY integrates a vectorized Sub-Graph encoder to capture environment context, a Variational Autoencoder to structure expert trajectories into a compact 32-dimensional latent manifold, and an ultra-lightweight MLP-Mixer decoder to eliminate quadratic attention complexity. Importantly, we introduce a latent-space drifting loss that shifts the complex distribution evolution entirely to the training phase. By formulating explicit attractive and repulsive forces, this mechanism empowers the model to synthesize novel, proactive maneuvers, such as active overtaking, that are virtually absent from the raw expert demonstrations. Extensive evaluations on the nuPlan benchmark demonstrate that MISTY achieves state-of-the-art results on the challenging Test14-hard split, with comprehensive scores of 80.32 and 82.21 in non-reactive and reactive settings, respectively. Operating at over 99 FPS with an end-to-end latency of 10.1 ms, MISTY offers an order-of-magnitude speedup over iterative diffusion planners while while achieving significantly robust generation.
\end{abstract}


\section{Introduction}
\IEEEPARstart{A}{s} the core component of autonomous driving systems, motion planning directly governs vehicle safety and ride comfort in complex, dynamic, multi-agent environments. The recent release of large-scale, high-fidelity closed-loop benchmarks such as nuPlan~\cite{nuplan} and the Waymo Open Motion Dataset (WOMD)~\cite{ettinger2021large} has accelerated a shift from traditional rule-based frameworks toward end-to-end imitation learning (IL). Deep neural planners (e.g., UniAD~\cite{hu2023planning} and PLUTO~\cite{cheng2024pluto}) have demonstrated the ability to handle interactive scenarios by distilling policies from vast expert demonstrations. However, human driving behavior is inherently non-deterministic and multi-modal: given identical histories and map topology, drivers may select different maneuvers that are all safe and rational. Models that rely on direct regression losses, such as Mean Squared Error (MSE), tend to average over these multiple modes, yielding kinematically infeasible trajectories. Accurately capturing and generating high-dimensional, continuous multi-modal distributions under strict onboard real-time constraints therefore remains a critical challenge.

To address the multi-modal generation bottleneck, advanced probabilistic frameworks, such as diffusion models and flow matching, have been widely adopted for trajectory planning~\cite{yang2024diffusiones}. By modeling the evolution from random noise to the manifold of expert trajectories, these methods produce high-quality, diverse candidate sets. However, standard diffusion models require multiple Neural Function Evaluations (NFEs) to solve the underlying Stochastic or Ordinary Differential Equations (SDEs/ODEs), incurring latency that is often unacceptable for closed-loop control. To overcome this, recent work has explored single-step generation. ConsistencyDrive~\cite{11288252} and Predictive Planner~\cite{li2025predictive} leverage Consistency Models~\cite{prasad2024consistency} for one-step inference, while MeanFuser~\cite{wang2026meanfuser} eliminates ODE numerical errors via mean flows. Despite lower latency, aggressively compressing the inference steps introduces truncation errors that, particularly in complex road geometries, cause ``feature drift'', where trajectories stray from drivable areas or collide with obstacles.

Mitigating feature drift and enforcing road compliance within a single-step framework requires a highly structured and physically meaningful representation space. We therefore adopt a hierarchical Sub-Graph architecture inspired by VectorNet~\cite{gao2020vectornet}, augmented with learnable attention pooling, to efficiently capture HD map constraints and dynamic agent interactions. Furthermore, we employ a lightweight Variational Autoencoder (VAE) to explicitly capture human driving intent and structure the trajectory feature space. By shifting from absolute coordinates to relative displacements and employing deep ResMLP networks, the VAE compresses high-dimensional expert trajectories into a compact, semantically discriminative 32-dimensional latent manifold. This structured space provides robust safety boundaries and topological priors for the subsequent generative model.

Building on the high-quality latent space extracted by the VAE, we introduce the recently proposed Drifting Models theory~\cite{deng2026generative} to effectively address the distribution shift problem in ultra-fast single-step generation. Unlike conventional generative models fitting velocity fields in low-dimensional coordinates, we shift the evolution of the pushforward distribution entirely to the training phase. Specifically, we formulate a Drifting Field Loss within the latent space. This loss formulates particle interactions under non-equilibrium dynamics: an attractive force draws generated feature vectors toward the manifold of expert driving, while a repulsive force among generated samples prevents mode collapse. For the decoder, we avoid the quadratic complexity of Transformer self-attention by adopting an ultra-lightweight MLP-Mixer architecture~\cite{tolstikhin2021mlp}. By alternating spatial and channel mixing with simple matrix multiplications, the MLP-Mixer scales linearly with sequence length, significantly reducing FLOPs and making it well-suited for the lightweight demands of onboard deployment.~\cite{feng2024road}

We evaluate MISTY extensively on the nuPlan closed-loop benchmark. Operating in a pure single-step inference mode, MISTY achieves competitive safety scores, matching or exceeding state-of-the-art generative planners on the challenging Test14-hard split (80.32 non-reactive, 82.21 reactive). At the same time, the full model runs at over 99 FPS with an end-to-end latency of 10.1 ms, representing an order-of-magnitude speedup over iterative diffusion methods. These results demonstrate that integrating vectorized scene encoding, a VAE-structured latent space, a lightweight MLP-Mixer decoder, and drifting model theory effectively bridges the gap between feature utilization and physical feasibility in single-step generative architectures. Our findings indicate that the drifting framework offers a viable and practical route toward real-time, multi-modal planning in complex urban environments.

The main contributions of this work are summarized as follows:
\begin{enumerate}
\item A latent-space drifting formulation that adapts Drifting Models to trajectory planning, shifting distribution transport to training and enabling single-step generation without iterative ODE solving.
\item A VAE-based trajectory manifold with relative displacement encoding, yielding a highly structured 32-dimensional latent space (with a 23$\times$ inter/intra-class ratio improvement over PCA) that facilitates explicit diversity control.
\item An ultra-lightweight MLP-Mixer decoder coupled with PCA reconstruction, which eliminates Transformer attention overhead and achieves 99 FPS inference with only 4.4 ms core generation latency.
\end{enumerate}

The remainder of this paper is organized as follows. Section~\ref{sec:related} reviews related work on generative trajectory planning and lightweight scene encoding architectures. Section~\ref{sec:method} details the proposed MISTY framework, including the vectorized encoder, the MLP-Mixer decoder, the VAE-based feature extraction pipeline, and the latent-space drifting loss formulation. Section~\ref{sec:experiments} presents extensive experimental validation on the nuPlan benchmark, covering feature space analysis, diversity evaluation, closed-loop planning performance, and inference efficiency comparisons. Finally, Section~\ref{sec:conclusion} concludes the paper and discusses directions for future work.

\section{Related Work}
\label{sec:related}
This section reviews the literature that underpins our framework. We first discuss the evolution of generative multi-modal planning, emphasizing the shift from iterative diffusion to single-step methods. We then survey advances in joint scene representation and lightweight decoding, targeting the bottleneck of efficient feature extraction and scalable computation in real-world autonomous driving.

\subsection{Generative Model-Based Multi-modal Trajectory Planning}
Traditional end-to-end planning often treats trajectory generation as deterministic regression, which is prone to mode collapse in interactive scenarios. Generative models have thus emerged as a powerful alternative due to their superior multi-modal distribution fitting. Early diffusion-based planners, such as Diffusion-ES~\cite{yang2024diffusiones} and Diffusion Planner~\cite{zheng2025diffusion}, generate high-quality candidates through Markovian denoising, using gradient-free evolutionary search or classifier guidance to enforce closed-loop safety. To reduce the inherent latency of iterative sampling, DiffusionDrive~\cite{liao2025diffusiondrive} introduces truncated diffusion initialized from clustered trajectory anchors, while DiffRefiner~\cite{zhou2025diff} employs a coarse-to-fine two-stage refinement guided by semantic interactions.

To further accelerate inference, flow matching techniques construct straight optimal transport paths. GoalFlow~\cite{xing2025goalflow} selects an optimal semantic goal to robustly condition a one-step Rectified Flow, and TrajFlow~\cite{yan2025trajflow} reformulates multi-agent prediction as highly efficient single-step masked spatiotemporal tensor reconstruction. Pushing toward ultra-fast deployment, Predictive Planner~\cite{li2025predictive} leverages consistency models guided by the Alternating Direction Method of Multipliers (ADMM), while MeanFuser~\cite{wang2026meanfuser} applies MeanFlow with continuous Gaussian mixture noise. However, aggressively compressing ODE solvers often amplifies discretization errors, causing kinematic drift under complex topologies. The Drifting Models theory~\cite{deng2026generative} structurally eliminates inference-time numerical integration by performing distribution transport during network updates, driven by explicit attractive and repulsive losses in feature space. Alternatively, frameworks like ResAD~\cite{zheng2025resad} decompose planning into a physics-based inertial reference and a generative residual, preserving generative capacity for behavioral uncertainty while mathematically guaranteeing baseline kinematic feasibility.

\subsection{Joint Scene Representation and Lightweight Decoding}
Efficiently modeling complex road geometry and multi-modal intent representations is critical under the strict computational budgets of vehicular edge devices. The representational paradigm has largely shifted from computationally expensive rasterized Bird's-Eye Views to sparse, vectorized formats. Pioneering architectures like VectorNet~\cite{gao2020vectornet} and LaneGCN~\cite{liang2020learning} model map elements and trajectories as polylines or dilated lane graphs to preserve topological connectivity. This was further advanced by VAD~\cite{jiang2023vad}, which introduced a fully vectorized, map-free end-to-end framework that explicitly enforces spatial constraints. To capture fine-grained local geometry alongside global context, contemporary approaches utilize hierarchical sub-graph encoders coupled with attention pooling mechanisms (e.g., HiVT~\cite{zhou2022hivt}), ensuring rotationally invariant and symmetric scene embeddings. Furthermore, to extract robust intent representations, Variational Autoencoders (VAEs)~\cite{chen2021trajvae} map raw trajectory coordinates into relative displacement manifolds, providing a semantically discriminative latent space that directly supports downstream generative distribution alignment.

Regarding decoding architectures, the quadratic computational complexity of Transformer multi-head self-attention imposes severe latency bottlenecks for long sequence modeling. Drawing inspiration from computer vision, recent trajectory planners have replaced heavy attention modules with MLP-Mixer~\cite{tolstikhin2021mlp} designs to achieve strict linear complexity. By alternating between spatial token mixing and channel feature mixing via basic matrix multiplications, models such as the lightweight Diffusion Mixer within PocketDP3~\cite{zhang2026pocketdp3} can process heterogeneous scene vectors utilizing a tiny fraction of the standard parameter count. When coupled with fine-grained trajectory tokenization, these lightweight decoders efficiently fuse temporal contexts without suffering from the compounding errors inherent to autoregressive generation, thereby enabling high-frequency, multi-modal generative planning in real-world deployments.

\begin{figure*}[!t]
\centering
\includegraphics[width=\textwidth]{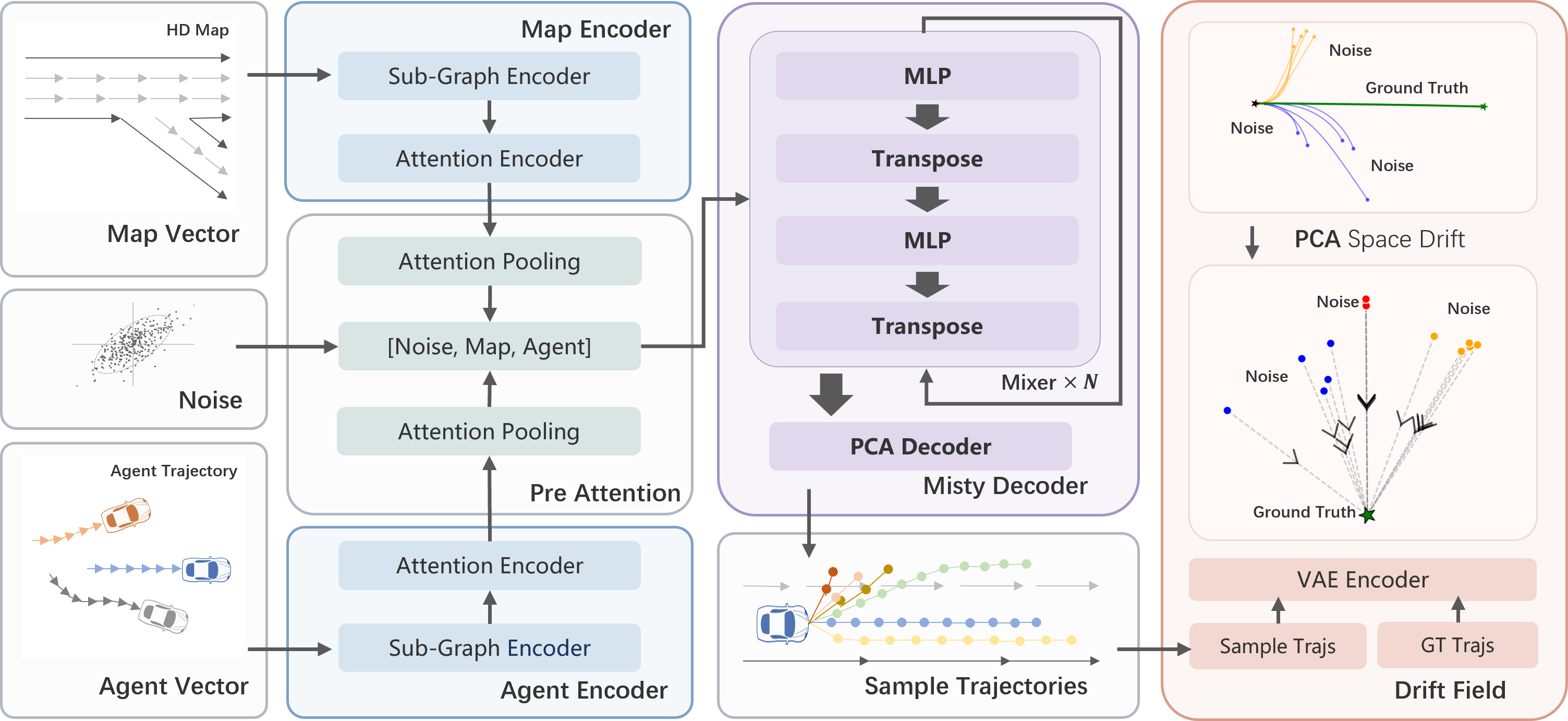}
\caption{The overall architecture of the MISTY framework, featuring a Vectorized Encoder, an MLP-Mixer-based Single-step Decoder, and a feature-space Drift Field loss mechanism.}
\label{fig1}
\end{figure*}

\section{Methodology}
\label{sec:method}
This section details the proposed MISTY framework. We first present an overview of the system architecture, followed by the specific designs of the vectorized Sub-Graph encoder, the MLP-Mixer-based drifting decoder, and the latent-space drifting loss.

\subsection{Overall Architecture}
This work proposes MISTY, a multi-modal trajectory planning model that operates directly on vectorized inputs: HD map data and historical agent trajectories. The architecture first extracts features using a Sub-Graph encoding approach inspired by VectorNet, producing representative map and agent embeddings. A concatenated vector comprising a noise token and the pooled agent and map features is then passed to the decoder. Drawing on the design principles of the Diffusion Transformer (DiT), we construct a Single-step Drifting Trajectory Decoder centered on an MLP-Mixer backbone, which works in synergy with a PCA Decoder to generate high-quality planning trajectories in a single forward pass. At the feature level, we apply drifting model theory to structure the distribution using a VAE encoder and compute the drifting loss directly in that latent space. The overall MISTY framework is illustrated in Fig. \ref{fig1}, and the specific functional modules are elaborated in the following subsections.

\subsection{Sub-Graph Encoder}
We represent road networks and agent histories in a vectorized format, where raw coordinates are transformed into structured directional vectors. Based on this representation, we design a Sub-Graph encoder consisting of three stages. First, a Multi-Layer Perceptron (MLP) projects the input channel information into a high-dimensional hidden space:
\begin{equation}
\mathbf{h}_{i,j} = \text{MLP}_{pt}(\mathbf{v}_{i,j}) + \text{Proj}(\mathbf{v}_{i,j})
\end{equation}
where $\mathbf{v}_{i,j}$ denotes the $j$-th vector of the $i$-th polyline and $\text{Proj}(\cdot)$ is a residual projection layer for feature alignment. Additional MLP layers then enhance the expressive power of the encoded features. To aggregate these point-level features into polyline-level representations, we implement a learnable attention pooling mechanism. The attention weight $\alpha_{i,j}$ and the resulting feature $\mathbf{f}_{i}$ are defined as follows:
\begin{equation}
\alpha_{i,j} = \frac{\exp(\mathbf{q}^\top \mathbf{W} \mathbf{h}_{i,j})}{\sum_{k} \exp(\mathbf{q}^\top \mathbf{W} \mathbf{h}_{i,k})}, \quad \mathbf{f}_{i} = \sum_{j} \alpha_{i,j} \mathbf{h}_{i,j}
\end{equation}
where $\mathbf{q}$ acts as a learnable query vector that prioritizes salient spatial or temporal information along the map length and agent timeline, respectively.

From these polyline features, we build separate map and agent encoders to capture the environmental context. The map encoder employs self-attention to model spatial constraints and lane connectivity, while the agent encoder captures dynamic interactions among traffic participants. After extracting these modal-specific features, the agent and map representations are concatenated and processed via global attention. This stage extracts cross-modal interactions, enabling the model to capture the dependencies between the ego vehicle's intent and the surrounding static and dynamic context.

The resulting unified global feature vector is then split into agent-specific and map-specific components. To prepare these high-dimensional representations for downstream generation, a second attention pooling stage extracts discrete agent and map tokens. This hierarchical encoding process is designed to extract heterogeneous scene semantics while preserving physical constraints, providing a robust contextual foundation for high-throughput, multi-modal trajectory generation.

\subsection{Single-step Drifting Decoder}
After obtaining the agent and map tokens, we concatenate them with a noise token $\boldsymbol{\epsilon} \sim \mathcal{N}(0, \mathbf{I})$ along the channel dimension. This fused representation, denoted as  $\mathbf{x}$, is then fed into the MLP-Mixer based decoder for trajectory inference. 

During decoding, we introduce a continuous parameter $\alpha$ as a conditional feature, replacing the discrete time step $t$ common in traditional diffusion models. Inspired by the conditioning mechanisms in DiT, we adaptively inject these conditional features into the MLP-Mixer blocks via adaptive layer normalization or affine transformations:
\begin{equation}
\mathbf{y} = \text{Mixer}(\mathbf{x} \mid \alpha, \mathbf{F}_{context})
\end{equation}
where $\mathbf{F}_{context}$ represents the encoded environmental tokens. This design allows the model to guide the generation process across different trajectory distributions. Through the stacked Mixer layers, which alternate between token-mixing and channel-mixing, the model effectively captures the deep correlations between the latent noise and the surrounding traffic context.

Finally, to map the processed latent features back into physical coordinates, we employ a PCA Decoder. Unlike standard regression heads, the PCA decoder reconstructs the full trajectory by projecting the high-dimensional mixer outputs $\mathbf{y}$ onto the principal component manifold, ensuring the generation of high-quality and kinematically plausible planning paths:

\begin{equation}
\mathcal{T} = \mathbf{W}_{pca} \cdot \mathbf{y} + \mu_{pca}
\end{equation}
where $\mathbf{W}_{pca}$ and $\mu_{pca}$ are the pre-trained principal components and mean trajectory, respectively. Leveraging the structural prior from PCA, this single-step inference pipeline significantly reduces computational latency while maintaining the expressive power of generative drifting models.

\subsection{Feature Space Extraction}
To model trajectory intent within a structured latent manifold, we design a feature extractor based on the encoding module of a Variational Autoencoder (VAE). A key step is transforming absolute coordinates into relative displacements. For a given raw expert trajectory $\tau$, we first compute its temporal displacements $\Delta \mathbf{p}_t = \mathbf{p}_t - \mathbf{p}_{t-1}$. This transformation explicitly removes absolute spatial biases, encouraging the network to focus strictly on intrinsic motion dynamics such as velocity profiles and heading variations.

The sequence of displacement features is then flattened and processed by a deep encoder to extract high-level semantic representations. The encoder uses a four-layer ResMLP (Residual Multi-Layer Perceptron) architecture with Pre-LayerNorm and GELU activations. This Pre-LayerNorm design effectively mitigates sensitivity to batch statistics, ensuring stable gradient flow and supporting deep hierarchical learning of complex trajectory patterns.

Through this deep encoding process, the network maps the high-dimensional motion features into the parameters of a latent probability distribution, specifically yielding the mean $\boldsymbol{\mu}$ and the standard deviation $\boldsymbol{\sigma}$. The final intent representation is then generated via the reparameterization trick:
\begin{equation}
\mathbf{z} = \boldsymbol{\mu} + \boldsymbol{\epsilon} \odot \boldsymbol{\sigma}, \quad \boldsymbol{\epsilon} \sim \mathcal{N}(0, \mathbf{I})
\end{equation}
where the resulting 32-dimensional latent vector $\mathbf{z}$ serves as a highly compressed, semantically discriminative representation of the underlying driving intent. By mapping raw physical trajectories into this structured latent manifold, we construct a compact and informative ``Drift Field'', providing essential feature-level conditioning to guide the subsequent generative drifting process.

\subsection{Latent-Space Drifting Loss Calculation}
Inspired by recent drifting models in image generation, we propose a latent-space drifting loss tailored specifically for trajectory planning. To construct the drifting field, we first prepare the target distributions. We cluster physically feasible trajectories into 16 distinct kinematic classes to serve as unconditional target samples, denoted as $\mathbf{z}_{unc}$. For a specific driving scenario, we utilize an compliance checker to identify valid, collision-free alternative trajectories (excluding the ground truth) as positive conditional samples $\mathbf{z}_{cond}$. In each training iteration, the model produces $K$ candidate trajectories that serve as negative samples $\mathbf{z}_{fake}$.

To ensure stable optimization across different latent dimensions, we apply dynamic feature normalization before computing the drift vectors. Using a distance-based interaction kernel, we compute the conditional drifting field $\mathbf{V}_{cond}$ and the unconditional drifting field $\mathbf{V}_{unc}$. Each field is constructed by calculating the attractive force from the target samples (positive/unconditional) and the repulsive force from the generated negative samples. To mitigate scale sensitivity, the attention weights for the drift vectors are computed using a bidirectional softmax normalization over the distance matrix.

Unlike image generation, which prioritizes pixel-level fidelity, autonomous driving demands diverse trajectory proposals to safely handle complex, interactive scenarios. To promote diversity, we interpolate the drifting fields using a continuous guidance scale $\alpha$. Crucially, we extend the range of $\alpha$ from the standard $[1.0, 4.0]$ to a broader interval of $[-0.5, 1.5]$:
\begin{equation}
\mathbf{V}_{total} = \mathbf{V}_{unc} + \alpha (\mathbf{V}_{cond} - \mathbf{V}_{unc})
\end{equation}
By allowing $\alpha$ to take negative values, the unconditional samples $\mathbf{z}_{unc}$ exert a distinct attractive pull, preventing the generated trajectories from collapsing into a single deterministic mode and ensuring rich multi-modality outputs.

Finally, we apply a magnitude normalization to the aggregated field to obtain $\bar{\mathbf{V}}_{total}$. During training, the physical trajectories generated by the PCA decoder are first projected into the VAE latent space to obtain $\bar{\mathbf{z}}_{fake}$. The training objective is then formulated as a Mean Squared Error (MSE) loss that drives these generated latent features toward their drifted targets, allowing gradients to backpropagate through the PCA decoder to the MLP-Mixer. To maintain stable training dynamics, a stop-gradient operator $\text{sg}(\cdot)$ is applied to the target state:
\begin{equation}
\mathcal{L}_{drift} = \frac{1}{K} \sum_{k=1}^{K} \|\bar{\mathbf{z}}_{fake}^{(k)} - \text{sg}(\bar{\mathbf{z}}_{fake}^{(k)} + \bar{\mathbf{V}}_{total}^{(k)})\|^2
\end{equation}
This formulation allows the network to iteratively shape the pushforward distribution of the planned trajectories within the structured latent manifold, aligning it with the target distribution specified by the guidance scale $\alpha$.

\section{Experiments}
\label{sec:experiments}
This section evaluates the MISTY framework extensively on the nuPlan dataset. The evaluation begins by detailing the dataset characteristics and the construction of unconditional samples. Subsequent experiments validate the structural superiority of the VAE-based feature space, analyze the training dynamics of the drifting model, and demonstrate the robust closed-loop planning performance compared against state-of-the-art baselines.

\subsection{Dataset and Data Preparation}
We train and evaluate our model on the widely adopted nuPlan dataset. As the first large-scale benchmark for closed-loop autonomous driving, nuPlan provides extensive real-world human driving data from diverse urban environments including Boston, Pittsburgh, Las Vegas, and Singapore. Unlike open-loop prediction datasets, nuPlan includes reactive agents and a high-fidelity closed-loop simulation engine, enabling accurate evaluation of long-term safety and kinematic feasibility in interactive settings.

To mitigate long-tail effects and balance behavior distributions, we perform stratified sampling based on nuPlan's rich scenario tags (e.g., intersections, lane changes, cut-ins). To construct the unconditional samples required for the drifting loss (Section III-E), we hierarchically cluster the extracted expert trajectories in physical coordinates. Specifically, we first cluster the massive expert trajectories into 10,000 fine-grained sub-classes, which are subsequently aggregated into 16 macroscopic behavioral categories with distinct kinematic semantics (e.g., cruising, left-turning, right-turning, and yielding). This hierarchical process establishes a structured dictionary of unconditional samples. During training, the model dynamically samples from this dictionary to obtain the unconditional samples ($\mathbf{z}_{unc}$) that represent the foundational manifold of driving behaviors, thereby providing a stable attractive reference for the distribution evolution in the drifting field.

\subsection{Feature Space Validation}
With the data dictionary in place, we pre-train the feature space extractor on the full nuPlan training split. To structure the latent manifold, we use the six primary macro-scenario tags (i.e., stationary, cruising, lane changing, left turning, right turning, and high-interaction) as auxiliary classification supervision. This objective encourages the encoder to capture high-level driving intents while compressing trajectories into a compact 32-dimensional space.

After 200 training epochs, we quantitatively evaluate the representation quality by measuring the distribution of the 16 unconditional dictionary classes across three distinct spaces: absolute physical coordinates, PCA-derived space, and our VAE latent space. As shown in Table \ref{tab1}, the absolute coordinate space exhibits a large inter-class distance, but this is mainly due to raw spatial scale rather than semantic separation, as evidenced by its high intra-class variance. In contrast, the PCA space fails to capture the complex non-linear dynamics of trajectories, yielding a low inter/intra ratio of 0.42. Our VAE latent space achieves a significantly superior inter/intra ratio of $23.64$. Crucially, it reduces the intra-class distance to $1.48$, demonstrating that the model successfully maps kinematically similar trajectories into highly compact clusters. This structural integrity is vital for the subsequent drifting process, as it provides a well-organized manifold that prevents distribution overlapping during generation.

\begin{figure}[htbp]
\centering
\includegraphics[width=3.4in]{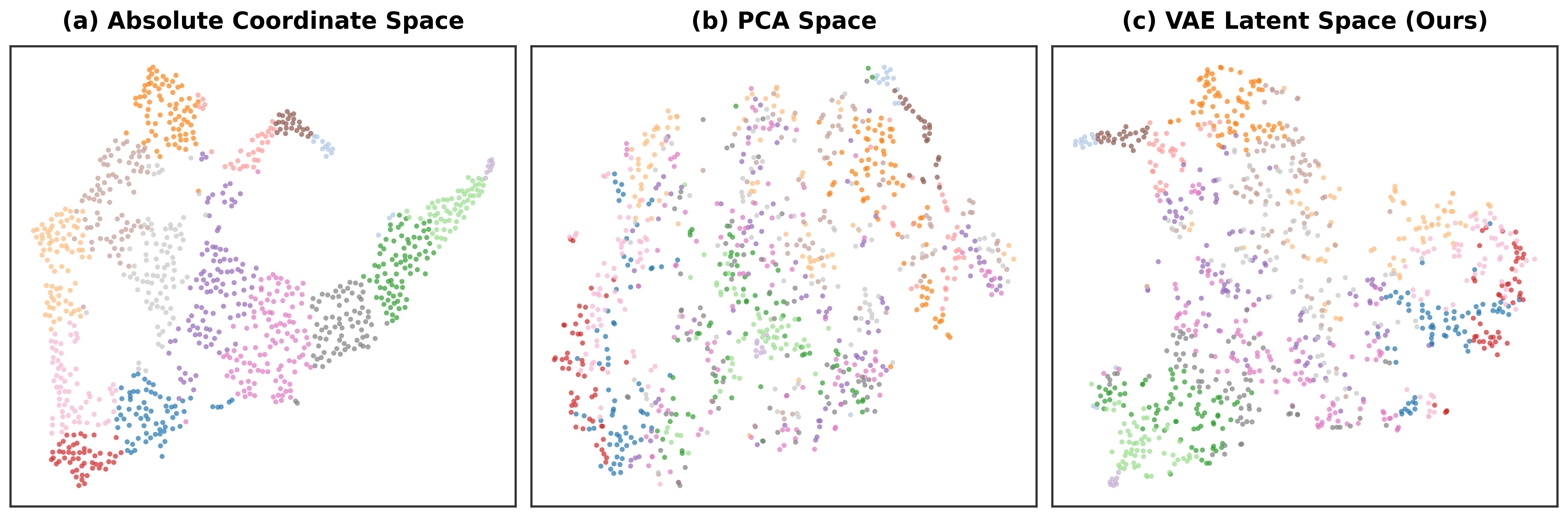}
\caption{Visualization of the 16 kinematic subclasses distribution across different spaces: (a) Absolute Coordinate Space, (b) PCA Space, and (c) VAE Latent Space. Each color corresponds to one of the 16 classes in the unconditional trajectory dictionary.}
\label{fig2}
\end{figure}

Furthermore, Fig. \ref{fig2} visualizes the feature distributions using t-SNE projection. In the visualization, each of the 16 distinct colors represents a specific kinematic subclass from the unconditional sample dictionary. As evident from Fig. \ref{fig2}(c), the VAE latent space forms a continuous yet clearly structured manifold. Unlike the overlapping clusters in PCA space, our VAE space facilitates distinct semantic separation while maintaining a dense distribution regulated by the KL-divergence. This highly discriminative feature space allows the drifting model to precisely fit the underlying expert distribution and generate physically consistent multimodal trajectories.

\begin{figure*}[!b]
\centering
\includegraphics[width=0.9\textwidth]{alpha_dynamics.png}
\caption{Visualization of the generated trajectories across varying guidance scales ($\alpha \in [-0.5, 1.5]$). Top Row: Physical space trajectories. Bottom Row: Corresponding VAE latent space distributions. Larger $\alpha$ values tightly concentrate samples around the ground truth (green star), while smaller $\alpha$ values drive samples to explore diverse unconditional anchors (purple crosses), ensuring rich multi-modality.}
\label{fig3}
\end{figure*}

\begin{table}[htbp]
\centering
\caption{Quantitative Comparison of Intra-class and Inter-class Distances for the 16 Unconditional Classes}
\label{tab1}
\resizebox{\columnwidth}{!}{
\renewcommand{\arraystretch}{1.2}
\begin{tabular}{lccc}
\hline\hline
Representation Space & Intra-class $\downarrow$ & Inter-class $\uparrow$ & Ratio (Inter/Intra) $\uparrow$ \\
\hline
Absolute Coordinate     & 50.51          & 228.64 & 4.63 \\
PCA Space (12-dim)      & 8.02           & 3.37   & 0.42 \\
VAE Latent Space (Ours) & \textbf{1.48}  & \textbf{34.93} & \textbf{23.64} \\
\hline\hline
\end{tabular}
}
\end{table}

Furthermore, Fig. \ref{fig2} visualizes the feature distributions using t-SNE projection. In the visualization, each of the 16 distinct colors represents a specific kinematic subclass from the unconditional sample dictionary. As evident from Fig. \ref{fig2}(c), the VAE latent space forms a continuous yet clearly structured manifold. Unlike the overlapping clusters in PCA space, our VAE space facilitates distinct semantic separation while maintaining a dense distribution regulated by the KL-divergence. This highly discriminative feature space allows the drifting model to precisely fit the underlying expert distribution and generate physically consistent multimodal trajectories.

\subsection{Drifting Guidance and Trajectory Diversity}
After validating the VAE feature space, we examine the training dynamics of the MISTY single-step drifting model. Fig. \ref{fig3} illustrates the effect of the guidance scale $\alpha$ on generated trajectories (top row) and latent distributions (bottom row). The visualizations align well with our theoretical formulation of the drifting field. At larger $\alpha$ (e.g., $\alpha \ge 1.0$), the generative distribution becomes tightly concentrated. In latent space, sampled points (orange) cluster tightly around the ground truth (green star). This strong conditional attraction ensures that the generated trajectories strictly adhere to the primary driving intent of the current scenario. In contrast, at lower or negative $\alpha$ (e.g., $\alpha \le 0.0$), the trajectories become more diverse. Driven by the inverted drifting field, the sampled points in latent space actively seek out and cluster around different unconditional anchors (purple crosses) rather than collapsing into a single mean state. This dynamic mechanism yields a highly diverse, multi-modal trajectory distribution, which is essential for providing varied, safe planning proposals to effectively handle complex corner cases.

\begin{table*}[!t]
\centering
\caption{Closed-Loop Planning Performance on the nuPlan Benchmark}
\label{tab2}
\renewcommand{\arraystretch}{1.2}

\newcolumntype{Z}{>{\hsize=1.6\hsize\raggedright\arraybackslash}X}
\newcolumntype{Y}{>{\hsize=0.85\hsize\centering\arraybackslash}X}

\begin{tabularx}{\textwidth}{ Z | Y Y | Y Y }
\hline
\multirow{2}{*}{Method} & \multicolumn{2}{c|}{Test14 (Score $\uparrow$)} & \multicolumn{2}{c}{Test14-hard (Score $\uparrow$)} \\
\cline{2-5}
 & NR & R & NR & R \\
\hline
\textit{Expert Baseline} & & & & \\
Log-replay & 94.03 & 75.86 & 85.96 & 68.80 \\
\hline
\textit{Rule-based / Hybrid} & & & & \\
IDM~\cite{treiber2000congested} & 70.39 & 74.42 & 56.15 & 62.26 \\
PDM-Closed~\cite{dauner2023parting} & 90.05 & \underline{91.63} & 65.08 & 75.19 \\
GameFormer~\cite{Huang_2023_ICCV} & 83.88 & 82.05 & 68.70 & 67.05 \\
PLUTO~\cite{cheng2024pluto} & \underline{92.23} & 90.29 & \underline{80.08} & 76.88 \\
Diffusion Planner w/ refine. ~\cite{zheng2025diffusion} & \textbf{94.80} & \textbf{91.75} & 78.87 & \underline{82.00} \\
MISTY (Ours) w/ refine & 91.24 & 90.70 & \textbf{80.32} & \textbf{82.21} \\
\hline
\textit{Learning-based / Generative} & & & & \\
PlanTF~\cite{cheng2024rethinking} & 85.62 & 79.58 & 69.70 & 61.61 \\
PLUTO w/o refine. ~\cite{cheng2024pluto} & 89.90 & 78.62 & 70.03 & 59.74 \\
Diffusion Planner~\cite{zheng2025diffusion} & 89.19 & 82.93 & 75.67 & 68.56 \\
MISTY (Ours) w/o refine & 78.88 & 82.25 & 70.64 & 75.82 \\
\hline
\multicolumn{5}{l}{\small \textit{NR: Non-reactive (log-replay); R: Reactive (IDM-based). Best results are in bold, second best are underlined.}} \\
\end{tabularx}
\end{table*}

\subsection{Closed-loop Planning Performance}
To evaluate the driving robustness of MISTY, we conduct extensive closed-loop simulations on the nuPlan benchmark. Following the standard evaluation protocol, we report the scores under two distinct settings: \textit{Non-Reactive} (NR) and \textit{Reactive} (R). In the NR setting, all other agents in the scene strictly execute their pre-recorded log trajectories regardless of the ego vehicle's actions. In contrast, the R setting utilizes an Intelligent Driving Model (IDM)~\cite{treiber2000congested} to simulate reactive behaviors, where other agents adjust their velocity and distance to respond to the ego vehicle's maneuvers. We evaluate our model across the \textit{Test14} and \textit{Test14-hard} splits, using the comprehensive nuPlan score as the primary metric.

To ensure safety and feasibility, we employ a streamlined selection pipeline to extract the optimal maneuver. First, a hard-constraint filter is applied to eliminate trajectory candidates that involve collisions or drivable area violations. Subsequently, inspired by the PDM-Closed~\cite{dauner2023parting}, we evaluate the remaining trajectories based on a composite metric of progress, safety, and comfort. This process ensures that the most robust and socially compliant trajectory is selected from MISTY's multi-modal proposals for closed-loop execution.

\begin{figure}[htbp]
\centering
\includegraphics[width=\columnwidth]{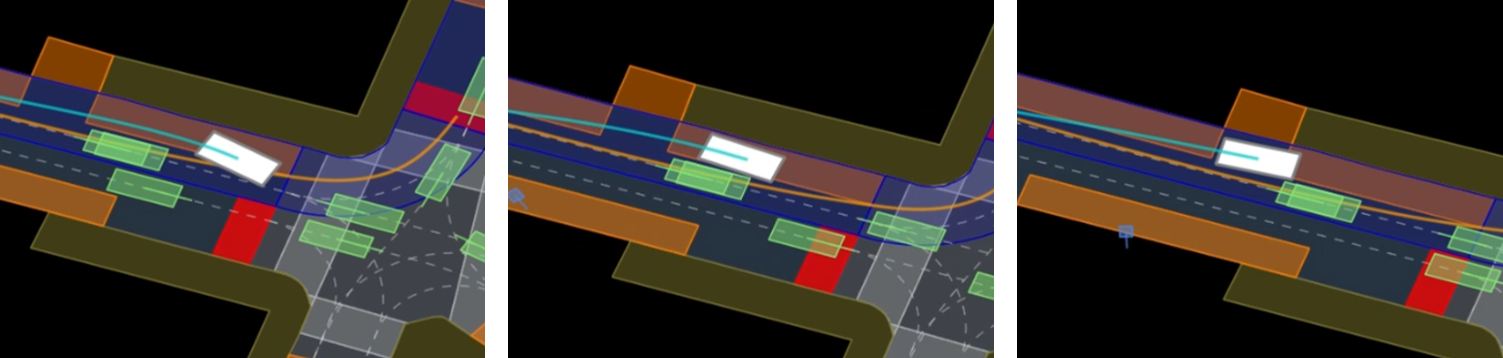}
\caption{Qualitative demonstration of MISTY's generalization capability. When the ego lane is blocked by a stationary vehicle, MISTY successfully generates and executes an active right-side overtaking maneuver, which represents an out-of-distribution behavior rarely found in the training dataset.}
\label{fig:overtake}
\end{figure}

As summarized in Table \ref{tab2}, MISTY demonstrates superior planning performance, particularly establishing a new state-of-the-art (SOTA) on the challenging \textit{Test14-hard} split. With the refinement module, MISTY achieves comprehensive scores of 80.32 and 82.21 in non-reactive and reactive settings, respectively, consistently surpassing all existing generative and hybrid baselines. This leading performance is fundamentally attributed to the model's robust capacity for generalization rather than mere dataset replication. Powered by the latent-space drifting mechanism, MISTY is capable of synthesizing novel behaviors that are virtually absent from the raw expert demonstrations. A representative case is illustrated in Fig. \ref{fig:overtake}, where MISTY successfully generates and executes an active right-side overtaking maneuver to bypass a lane blockage, whereas traditional models often revert to passive car-following or stall. This ability to explore diverse, proactive, and kinematically safe solutions within the structured latent manifold effectively prevents mode collapse, representing the decisive factor in achieving state-of-the-art robustness across complex and interactive scenarios.

While MISTY establishes a clear advantage in safety-critical situations, we observe slightly lower scores in the nominal \textit{Test14} split compared to Diffusion Planner, and the raw generative model (\textit{w/o refine}) exhibits a performance drop. This can be attributed to two main factors. First, the inherent trajectory smoothness of single-step generation is still slightly inferior to that of multi-step iterative refinement, incurring minor penalties in the comfort sub-metric. Second, the training dataset exhibits a long-tail distribution with extreme sparsity of high-quality emergency braking or precise stopping maneuvers. Consequently, the single-step model struggles to perfectly learn these deceleration profiles natively, occasionally leading to slight instability when approaching a stop line. However, our lightweight refinement module effectively resolves these artifacts, significantly recovering the performance. Ultimately, coupling its SOTA interactive safety with an extreme 99 FPS inference speed, MISTY offers a superior and practical solution for real-world onboard deployment.

\subsection{Inference Efficiency Analysis}
A critical bottleneck in deploying generative planners for autonomous driving is inference latency, which typically stems from heavy multi-head attention and iterative multi-step ODE solvers. Table \ref{tab3} presents a quantitative comparison of inference throughput among recent learning-based planners. By offloading the complex distribution evolution to the training phase via the feature-space drifting field, MISTY fundamentally eliminates the need for iterative sampling, enabling high-fidelity \textit{single-step} generation during inference.

Comprehensive profiling on an NVIDIA RTX 4090 GPU reveals that the full end-to-end model inference (encompassing VectorNet-based high-dimensional scene encoding and trajectory generation) requires only 10.1 ms (approximately 99 FPS) after excluding outlier spikes via a trimmed mean approach. Furthermore, the core trajectory generation process itself, executed by our MLP-Mixer-based decoder, consumes a mere 4.4 ms to yield 1024 multi-modal trajectory proposals in a single forward pass. While traditional multi-step generative planners (e.g., Diffusion Planner) suffer from severe inference latency, MISTY strikes an optimal structural balance. It retains a powerful, context-rich scene encoder to ensure complex scenario understanding, while optimizing the generative decoder to maximize computational efficiency. This low latency guarantees real-time, high-throughput responsiveness, making MISTY highly practical for on-board deployment.

\begin{table}[htbp]
\centering
\caption{Inference Efficiency Comparison of Generative Planners}
\label{tab3}
\resizebox{\columnwidth}{!}{
\renewcommand{\arraystretch}{1.2}
\begin{tabular}{lcc}
\hline\hline
Method & Decoding Steps & Throughput (FPS $\uparrow$) \\
\hline
Diffusion Planner~\cite{zheng2025diffusion} & 10  & $\sim$13 \\
PLUTO~\cite{cheng2024pluto}                 & -   & $\sim$47 \\
PlanTF~\cite{cheng2024rethinking}           & -   & $\sim$97 \\
MISTY (Ours)                                & 1   & \textbf{$\sim$99} \\
\hline\hline
\end{tabular}
}
\end{table}

\section{Conclusion}
\label{sec:conclusion}
This paper presented MISTY, a high-throughput single-step generative motion planner designed to address the inference latency bottleneck of diffusion-based architectures in autonomous driving. By integrating a vectorized scene encoder, a VAE-derived structured latent manifold, and an ultra-lightweight MLP-Mixer decoder, MISTY shifts the complex multi-modal distribution evolution to the training phase via a novel latent-space drifting loss. This formulation effectively balances attractive forces toward the expert manifold with repulsive forces among generated samples, eliminating the discretization errors common in single-step generation while preserving essential trajectory diversity.

Extensive closed-loop evaluations on the nuPlan benchmark demonstrate that MISTY achieves state-of-the-art safety and interactive performance. Notably, it surpasses the prior best generative baseline on the challenging Test14-hard reactive split, achieving a top score of 82.21. This leading performance is primarily attributed to the model's robust capacity for generalization, which enables the synthesis of proactive and out-of-distribution behaviors, such as active overtaking, that are rarely observed in the raw expert dataset. Such ability to discover novel yet kinematically safe solutions is fundamental for navigating complex, high-interaction corner cases.

With an end-to-end inference latency of 10.1 ms ($\sim$99 FPS) and a core generation time of only 4.4 ms, MISTY satisfies the stringent real-time requirements of vehicular edge devices. These results indicate that generative trajectory planning can achieve high-fidelity multi-modal outputs without sacrificing computational efficiency or physical feasibility. Consequently, the drifting-based framework offers a highly practical and scalable solution for real-world autonomous driving deployments in complex urban environments.

\bibliographystyle{IEEEtran}
\bibliography{references}

\end{document}